\documentclass[sigconf, screen]{acmart}
\AtBeginDocument{%
  }

\setcopyright{acmlicensed}
\usepackage{pgfplots}
\usepackage{pifont}

\usepackage{amssymb}
\usepackage{multirow}
\usepackage{hyperref}
\usepackage{bbm}
\usepackage{subcaption} 
\usepackage{xcolor}    
\pgfplotsset{compat=1.18}
\usepackage{tikz}
\copyrightyear{2026}
\acmYear{2026}
\acmDOI{XXXXXXX.XXXXXXX}
\acmConference[MM '26]{Proceedings of the 34th ACM International Conference on Multimedia}{November 10--14, 2026}{Rio de Janeiro, Brazil}
\acmISBN{XXX-X-XXXX-XXXX-X/2026/XX}
\usepackage{booktabs}

\usepackage[table]{xcolor}
\newlength{\methodcolwd}
\begin{document}

\title{LEGO: LoRA-Enabled Generator-Oriented Framework for Synthetic Image Detection}
\acmSubmissionID{5440}
\author{Yutong Xiao}
\affiliation{%
	\institution{University of Electronic Science and Technology of China}
	\city{Chengdu}
	\country{China}
}

\author{Ran Ran}
\affiliation{%
	\institution{University of Electronic Science and Technology of China}
	\city{Chengdu}
	\country{China}
}

\author{Jiwei Wei}
\authornote{Corresponding author.Email:{mathematic6@gmail.com} (Jiwei Wei)}
\affiliation{%
	\institution{University of Electronic Science and Technology of China}
	\city{Chengdu}
	\country{China}
}

\author{Shuchang Zhou}
\affiliation{%
	\institution{University of Electronic Science and Technology of China}
	\city{Chengdu}
	\country{China}
}

\author{Ke Liu}
\affiliation{%
	\institution{University of Electronic Science and Technology of China}
	\city{Chengdu}
	\country{China}
}

\author{Zheng Ziqiang}
\affiliation{%
	\institution{University of Electronic Science and Technology of China}
	\city{Chengdu}
	\country{China}
}

\author{Caiyan Qin}
\affiliation{%
	\institution{Harbin Institute of Technology, Shenzhen}
	\city{Shenzhen}
	\country{China}
}

\begin{abstract}
The rapid advancement of generative technologies has made synthetic images nearly indistinguishable from real ones, thereby creating an urgent need for robust detectors to counter misinformation. However, existing methods mainly rely on universal artifact features that are shared across multiple generators. We observe that as the diversity of generators increases, the overlap of these common features gradually decreases. This severely undermines model generalization. In contrast, focusing only on unique artifacts tends to cause overfitting to specific forgery patterns.
To address this challenge, we propose \textbf{LEGO} (\textbf{L}oRA-\textbf{E}nabled \textbf{G}enerator-\textbf{O}riented Framework). The core mechanism of LEGO employs an MLP to modulate multiple LoRA (Low-Rank Adaptation) blocks, each pre-trained to capture the unique artifacts of a specific generator, followed by attention-based feature fusion. Unlike conventional methods that seek a single universal solution, LEGO delegates unique artifact extraction to specialized LoRA modules by dividing its training procedure into two stages. Each LoRA module is individually trained on a single-generator dataset to learn generator-specific representations, then MLP and attention layers are trained on mixed datasets to dynamically regulate the contribution of each module. Benefiting from its modular yet robust design, LEGO can be naturally extended by incorporating new LoRA modules for adaptation to newly emerging next-generation datasets, while still achieving substantially better performance than prior SOTA methods with fewer than {30,000} training images, less than 10\% of their training data, and only 5 epochs in each training stage.Project page: \href{https://github.com/seanstep/LEGO_LoRA_Enabled_Generator_Oriented_Framework}{LEGO: LoRA-Enabled Generator-Oriented Framework}

\end{abstract}

\ccsdesc[500]{Computing methodologies~Visual inspection}
\ccsdesc[500]{Computing methodologies~Image representations}
\ccsdesc[500]{Computing methodologies~Visual content-based indexing and retrieval}

\keywords{Deepfake Detection, Fake Image Classification, Security, Low-Rank Adaptation, Gating Mechanism}

\maketitle
\section{Introduction}
With the rapid development of generative models, producing highly realistic synthetic images has become increasingly easy~\cite{goodfellow2014generative,karras2019style,ho2020denoising,song2020denoising,dhariwal2021diffusion,rombach2022high,ramesh2022hierarchical,zhang2023adding}, which greatly complicates the verification of digital information. When maliciously used to spread misinformation or manipulate public opinion, such content can severely undermine trust in the modern digital ecosystem. This challenge is further amplified by the rapid emergence of diverse generators and increasingly realistic synthesis pipelines. Therefore, developing robust and reliable AI-generated image detectors has become an urgent research problem.

Early studies on AI-generated image detection mainly aim to identify transferable artifacts left by generative models, exploring cues from spatial textures, frequency patterns, and cross-model inconsistencies~\cite{wang2020cnn,frank2020leveraging,liu2021spatial,ju2022fusing,liu2022detecting,tan2023learning,ojha2023towards}. More recent methods have further extended this line of research by exploiting diffusion reconstruction inconsistency, intrinsic-dimensionality and inversion-based clues for diffusion images, representation priors from large vision-language models, image transformation strategies, frequency-aware modeling, and bias-reduced training paradigms~\cite{wang2023dire,lorenz2023detecting,cazenavette2024fakeinversion,cozzolino2024raising,li2025improving,tan2024frequency,yan2026dual,guillaro2025bias}. In particular, CLIP-based methods have shown promising generalization ability by leveraging strong semantic priors~\cite{radford2021learning,cozzolino2024raising,wu2025generalizable,tan2025c2p}, while preprocessing- and representation-based methods attempt to amplify synthetic traces before classification~\cite{cavia2024real,li2025improving,zhong2023patchcraft,sinitsa2024deep}. However, these approaches still rely, either explicitly or implicitly, on the assumption that different generators share sufficiently common artifacts, which may become increasingly weak as generative models continue to evolve.

\begin{figure*}[t]
    \centering
	\def\SubFigW{0.49\textwidth}      
 \begin{subfigure}[t]{0.48\textwidth}
    \vspace{0pt}
    \centering

    \def\xlego{0.30}
    \def\xaide{3.24}
    \def\xnpr{3.24}
    \def\xeffort{7.20}
    \def\xtridetect{3.24}
    \def\xhidanet{25.81}

    \def\ylego{83.60}
    \def\yaide{62.56}
    \def\ynpr{58.05}
    \def\yeffort{64.59}
    \def\ytridetect{67.88}
    \def\yhidanet{79.10}

    \begin{tikzpicture}
    \begin{axis}[
        width=0.8\linewidth,
        height=0.8\linewidth,
        xmode=log,
        log basis x={10},
        xmin=0.22, xmax=30,
        ymin=56, ymax=85,
        xlabel={\textbf{Training Images ($\times 10^5$)}},
        ylabel={\textbf{ACC on Chameleon (\%)}},
        title={\textbf{Efficiency--Performance Trade-off}},
        title style={font=\small, yshift=-2pt},
        xtick={0.3,1,3,10,25.81},
        xticklabels={0.3,1,3,10,25.8},
        ytick={58,62,66,70,74,78,82},
        tick label style={font=\scriptsize},
        label style={font=\scriptsize},
        axis line style={black, line width=0.9pt},
        tick style={black, line width=0.9pt},
        grid=major,
        grid style={line width=0.35pt, draw=gray!25},
        legend style={
            font=\tiny,
            draw=gray!45,
            fill=white,
            rounded corners=1pt,
            at={(0.03,0.05)},
            anchor=south west
        },
        legend cell align=left,
        clip=false
    ]

    \addplot[dashed, gray!70, line width=0.8pt, forget plot]
        coordinates {(\xlego,56) (\xlego,85)};
    \addplot[dashed, gray!70, line width=0.8pt, forget plot]
        coordinates {(0.22,\ylego) (30,\ylego)};

    \addplot[
        only marks,
        mark=*,
        mark size=4.4pt,
        color=cyan!85!blue,
        forget plot
    ] coordinates {(\xaide,\yaide)};

    \addplot[
        only marks,
        mark=square*,
        mark size=4.4pt,
        color=green!70!black,
        forget plot
    ] coordinates {(\xnpr,\ynpr)};

    \addplot[
        only marks,
        mark=diamond*,
        mark size=4.6pt,
        color=magenta!80!black,
        forget plot
    ] coordinates {(\xeffort,\yeffort)};

    \addplot[
        only marks,
        mark=triangle*,
        mark size=4.8pt,
        color=orange!95!black,
        forget plot
    ] coordinates {(\xtridetect,\ytridetect)};

    \addplot[
        only marks,
        mark=pentagon*,
        mark size=5.6pt,
        color=blue!70!black,
        forget plot
    ] coordinates {(\xhidanet,\yhidanet)};

    \addplot[
        only marks,
        mark=+,
        mark size=5.4pt,
        line width=2.0pt,
        color=red!90!black,
        forget plot
    ] coordinates {(\xlego,\ylego)};

    \addlegendimage{only marks, mark=*, mark size=4.4pt, color=cyan!85!blue}
    \addlegendentry{AIDE}

    \addlegendimage{only marks, mark=square*, mark size=4.4pt, color=green!70!black}
    \addlegendentry{NPR}

    \addlegendimage{only marks, mark=diamond*, mark size=4.6pt, color=magenta!80!black}
    \addlegendentry{Effort}

    \addlegendimage{only marks, mark=triangle*, mark size=4.8pt, color=orange!95!black}
    \addlegendentry{TriDetect}

    \addlegendimage{only marks, mark=pentagon*, mark size=5.6pt, color=blue!70!black}
    \addlegendentry{HiDA-Net}

    \addlegendimage{only marks, mark=+, mark size=5.4pt, line width=2.0pt, color=red!90!black}
    \addlegendentry{Ours}

    \node[font=\tiny, anchor=west] at (axis cs:\xaide*1.03,\yaide+0.25) {AIDE};
    \node[font=\tiny, anchor=west] at (axis cs:\xnpr*1.03,\ynpr-0.25) {NPR};
    \node[font=\tiny, anchor=west] at (axis cs:\xeffort*1.02,\yeffort+0.20) {Effort};
    \node[font=\tiny, anchor=west] at (axis cs:\xtridetect*1.03,\ytridetect+0.20) {TriDetect};
    \node[font=\tiny, anchor=east] at (axis cs:\xhidanet/1.20,\yhidanet+0.15) {HiDA-Net};
    \node[font=\tiny\bfseries, text=red!90!black, anchor=west] at (axis cs:\xlego*1.06,\ylego+0.35) {Ours};

    \end{axis}
    \end{tikzpicture}
    \caption{Training-Data Efficiency vs. Performance.}
    \label{fig:efficiency_performance}
\end{subfigure}
    \hfill
    \begin{subfigure}[t]{0.49\textwidth}
        \vspace{0pt}
        \centering
        \begin{tikzpicture}
        \begin{axis}[
            width=0.8\linewidth,
            height=0.8\linewidth,
            xmin=14.5, xmax=18.6,
            ymin=65, ymax=81,
            xlabel={\textbf{Std. across datasets}},
            ylabel={\textbf{Mean Acc on AIGIBench}},
            title={\textbf{Balanced Generalization}},
            title style={font=\small, yshift=-2pt},
            xtick={15,16,17,18},
            ytick={66,68,70,72,74,76,78,80},
            tick label style={font=\scriptsize},
            label style={font=\scriptsize},
            axis line style={black, line width=0.9pt},
            tick style={black, line width=0.9pt},
            grid=major,
            grid style={line width=0.35pt, draw=gray!25},
            legend style={
                font=\tiny,
                draw=gray!45,
                fill=white,
                rounded corners=1pt,
                at={(0.03,0.05)},
                anchor=south west
            },
            legend cell align=left,
            clip=false
        ]

        \addplot[dashed, gray!70, line width=0.8pt, forget plot]
            coordinates {(14.87,65) (14.87,81)};
        \addplot[dashed, gray!70, line width=0.8pt, forget plot]
            coordinates {(14.5,80.20) (18.6,80.20)};

        \addplot[only marks, mark=*, mark size=4.2pt, color=cyan!85!blue, forget plot]
            coordinates {(18.04,76.79)};
        \addplot[only marks, mark=triangle*, mark size=4.5pt, color=orange!95!black, forget plot]
            coordinates {(16.22,71.47)};
        \addplot[only marks, mark=diamond*, mark size=4.5pt, color=magenta!80!black, forget plot]
            coordinates {(17.98,74.13)};
        \addplot[only marks, mark=square*, mark size=4.2pt, color=green!70!black, forget plot]
            coordinates {(17.68,66.59)};
        \addplot[only marks, mark=pentagon*, mark size=4.5pt, color=yellow!85!orange, forget plot]
            coordinates {(17.50,69.97)};
        \addplot[only marks, mark=+, mark size=5.0pt, line width=1.8pt, color=red!90!black, forget plot]
            coordinates {(14.87,80.20)};

        \addlegendimage{only marks, mark=*, mark size=4.2pt, color=cyan!85!blue}
        \addlegendentry{AIDE}
        \addlegendimage{only marks, mark=triangle*, mark size=4.5pt, color=orange!95!black}
        \addlegendentry{CO-SPY}
        \addlegendimage{only marks, mark=diamond*, mark size=4.5pt, color=magenta!80!black}
        \addlegendentry{Effort}
        \addlegendimage{only marks, mark=square*, mark size=4.2pt, color=green!70!black}
        \addlegendentry{NPR}
        \addlegendimage{only marks, mark=pentagon*, mark size=4.5pt, color=yellow!85!orange}
        \addlegendentry{DFFreq}
        \addlegendimage{only marks, mark=+, mark size=5.0pt, line width=1.8pt, color=red!90!black}
        \addlegendentry{Ours}

        \node[font=\tiny, anchor=west] at (axis cs:18.09,76.95) {AIDE};
        \node[font=\tiny, anchor=west] at (axis cs:16.31,71.30) {CO-SPY};
        \node[font=\tiny, anchor=west] at (axis cs:18.07,74.00) {Effort};
        \node[font=\tiny, anchor=west] at (axis cs:17.77,66.42) {NPR};
        \node[font=\tiny, anchor=west] at (axis cs:17.58,69.82) {DFFreq};
        \node[font=\tiny\bfseries, text=red!90!black, anchor=west] at (axis cs:14.98,80.38) {Ours};

        \end{axis}
        \end{tikzpicture}
        \caption{Mean--standard-deviation comparison across AIGIBench subsets.}
        \label{fig:mean_variance}
    \end{subfigure}
	\vspace{-3pt}
		\caption{Comparison of detector efficiency and generalization. Left: data-efficiency--performance trade-off using ACC on Chameleon. Right: mean--standard-deviation comparison across AIGIBench subsets.}
    \label{fig:combined_aigibench}
    \vspace{-0pt}
\end{figure*}
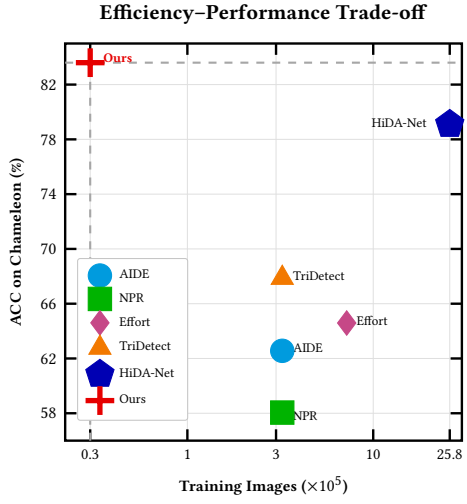
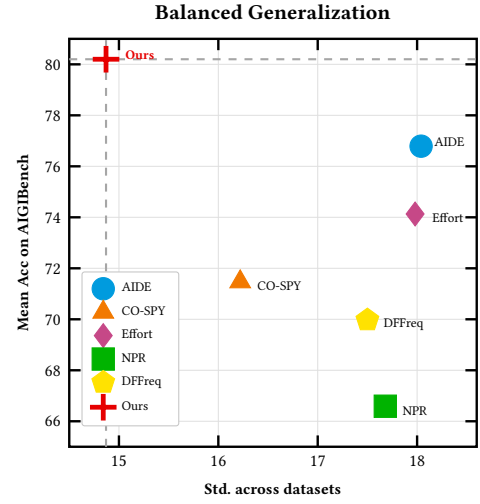

Recent studies further highlight the difficulty of robust real-world generalization. Large-scale benchmarks and more realistic evaluation protocols have already raised the bar for this task~\cite{zhu2023genimage,hong2024wildfake,pal2024semi,pellegrini2025ai}. In particular, AIDE~\cite{yan2024sanity} shows that many existing detectors perform well on conventional benchmarks but degrade substantially on more challenging in-the-wild settings. This observation is further supported by AIGIBench, recent generator-diversity studies, and semantic-bias analyses, which together suggest that AI-generated image detection is far from being a solved problem under realistic and diverse distributions~\cite{li2025artificial,park2025community,zheng2024breaking}. To address this challenge, several recent methods have explored new directions. Effort improves generalized detection through efficient orthogonal artifact modeling~\cite{yan2024effort}. TriDetect improves generalized detection by modeling latent structural patterns within fake images beyond simple binary supervision~\cite{nguyen2026beyond}. HiDA-Net emphasizes the importance of preserving native-resolution details for robust high-resolution AIGC detection~\cite{mu2025no}. SimLBR instead learns a tighter boundary around real images to improve robustness under distribution shifts~\cite{dhakal2026simlbr}. SPAI enhances robustness across different image resolutions through spectral learning~\cite{karageorgiou2025any}, while DFFreq further exploits frequency-aware modeling for synthetic image detection~\cite{yan2026dual}. Other recent efforts also improve cross-generator robustness from complementary perspectives, including image-transformation-based generalization, bias-free training, and language-guided contrastive learning~\cite{li2025improving,guillaro2025bias,wu2025generalizable}. Although these methods achieve promising progress, their performance on challenging benchmarks indicates that generalizable detection remains an open problem.

To address this issue, we argue that pursuing a single ``universal artifact'' is ultimately insufficient. Instead, detectors should explicitly model generator-oriented unique artifacts, so that they can remain discriminative on seen generators while adapting efficiently to newly emerging ones. This view is also supported by recent findings on large generator diversity and semantic shortcut bias in AI-generated image detection~\cite{park2025community,zheng2024breaking,li2025artificial}. Based on this insight, we propose \textbf{LEGO}, a \textbf{L}oRA-\textbf{E}nabled \textbf{G}enerator-\textbf{O}riented framework for synthetic image detection. LEGO employs a pre-trained OpenCLIP backbone together with multiple generator-specific LoRA modules~\cite{radford2021learning,hu2022lora}, an MLP-based routing mechanism, and multi-level feature fusion to learn discriminative artifact representations in a modular manner. Similar to assembling LEGO bricks, the framework can be flexibly extended by attaching new LoRA modules for new generators, enabling efficient adaptation without retraining the entire model.

Extensive experiments on challenging benchmarks demonstrate the effectiveness of LEGO. Compared with recent strong baselines, including Effort, TriDetect, HiDA-Net, and SimLBR~\cite{yan2024effort,nguyen2026beyond,mu2025no,dhakal2026simlbr}, LEGO achieves consistently stronger performance in challenging in-the-wild settings, while requiring only a subset of the training data, further demonstrating its efficiency and practicality. As illustrated in Figure~\ref{fig:combined_aigibench}, despite being trained on substantially fewer seen generator subsets than competing methods, LEGO achieves higher average accuracy with lower variance across fake images by different generators. This suggests that LEGO delivers consistently strong performance across diverse datasets, rather than relying on a few exceptionally high scores that may mask a lack of discriminative capability on specific benchmarks.

Our contributions are summarized as follows:
\begin{itemize}
    \item We propose a novel paradigm for AI-generated image detection, showing that capturing generator-oriented unique artifacts is more effective than seeking universal ones in complex and diverse scenarios.
    
    \item We design a scalable framework that can be extended like LEGO bricks by adding new LoRA modules for newly emerging generators, enabling rapid adaptation with substantially reduced retraining and computational cost.
    
    \item Our framework {LEGO} achieves substantially better performance than recent strong baselines while using much less training data and only a few epochs of training, demonstrating superior effectiveness under a more economical training setting.
\end{itemize}

\begin{figure*}
	\centering
	\includegraphics[width=0.86\linewidth]{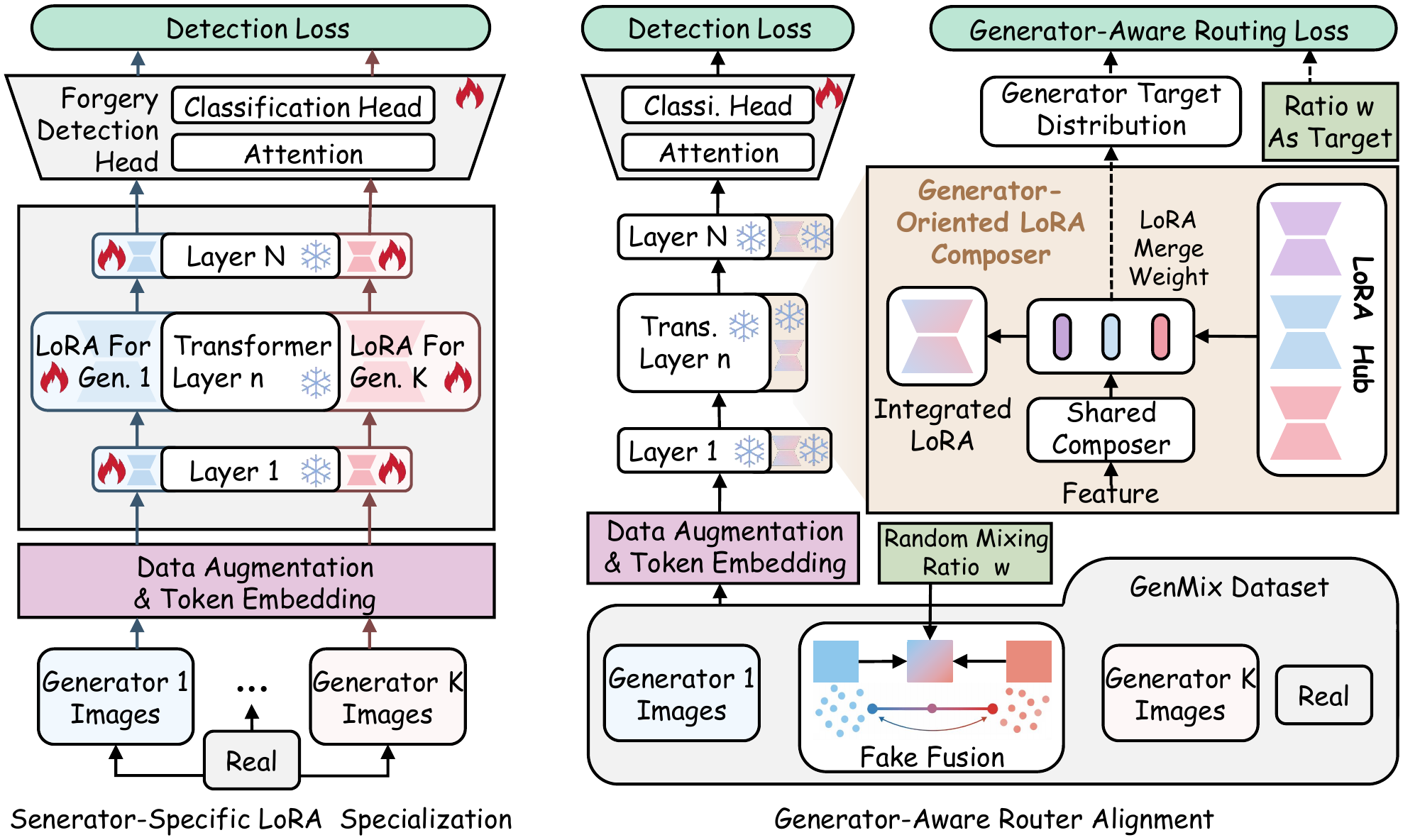}
	\vspace{-0pt}
	\caption{Overview of the proposed LEGO framework. LEGO is built upon a frozen CLIP backbone and trained in two stages. In Generator-Specific LoRA Specialization, each LoRA branch is individually optimized on a single-generator subset to capture distinctive artifact patterns. In Generator-Aware Router Alignment, the specialized branches are frozen and integrated into a LoRA Hub, while a router learns to predict the LoRA Merge Weight under mixed-generator training. The resulting Integrated LoRA is injected into the backbone, and the representation is refined by the Forgery Detection Head for prediction.}
	\label{fig:framework}
	\vspace{-0pt}
\end{figure*}

\section{Related Work}
\noindent\textbf{Synthetic image detection.}
With the rapid development of modern image generators, including GAN-based models~\cite{goodfellow2014generative,karras2017progressive,karras2019style} and diffusion-based models~\cite{ho2020denoising,song2020denoising,dhariwal2021diffusion,rombach2022high,ramesh2022hierarchical,zhang2023adding}, synthetic image detection has become increasingly challenging. Early detectors mainly rely on discriminative artifacts left by image generators. Representative works show that synthetic images can be exposed through visual artifacts in spatial textures, frequency patterns, and cross-model inconsistencies~\cite{wang2020cnn,frank2020leveraging,liu2020global,liu2021spatial,ju2022fusing,liu2022detecting}. In the broader forensic literature, related ideas have also been explored through consistency-aware representation learning and transformer-based inconsistency modeling~\cite{ni2022core,zhuang2022uia}. Later studies further investigate gradient-based, local texture-patch, and fingerprint-style forensic cues~\cite{tan2023learning,zhong2023patchcraft,sinitsa2024deep}. As diffusion models become dominant, DIRE~\cite{wang2023dire}, local intrinsic dimensionality~\cite{lorenz2023detecting}, and inversion-based detection~\cite{cazenavette2024fakeinversion} further show that diffusion-generated images exhibit distinctive reconstruction and representation behaviors that can be exploited for detection. More recent studies also revisit practical deployment settings, such as CLIP-based forensic detection~\cite{cozzolino2024raising}, real-world real-time detection~\cite{cavia2024real}, local pixel dependency modeling~\cite{liang2025ferretnet}, and semantic--pixel joint reasoning~\cite{cheng2025co}. Despite promising progress, these methods are sensitive to generator shifts and tend to generalize poorly when the testing distribution differs substantially from the training source.

\noindent\textbf{Frequency-aware and generalization-oriented methods.} To improve transferability, recent studies move beyond pixel-space classification and instead exploit transformed feature spaces, architectural priors, and generator diversity. Frequency-aware learning has proven effective for capturing subtle synthetic traces~\cite{tan2024frequency,yan2026dual}, while revisiting generator up-sampling operations provides useful inductive bias for generalized detection~\cite{tan2024rethinking}. Beyond representation design, several works explicitly target robustness and generalization. Early studies explored universal detectors trained across generators~\cite{ojha2023towards}, while SAFE~\cite{li2025improving} improves detection through image transformation and bias-free training reduces shortcut learning~\cite{guillaro2025bias}. Community Forensics~\cite{park2025community} further shows that large generator diversity is crucial for improving performance on unseen models. More recent efforts investigate semantic bias, disentangled artifact learning, and information-bottleneck-based universal detection~\cite{zheng2024breaking,yan2024effort,zhang2025towards}. Zero-shot and open-set settings have also attracted increasing attention, including forensic self-descriptions for source attribution~\cite{nguyen2025forensic} and multimodal large-language-model-based explainable detection~\cite{zhou2025aigi}. 
Meanwhile, benchmark studies consistently show that generalized detection remains far from solved. GenImage, WildFake, Semi-Truths, AI-GenBench, AIGIBench, and MIRAGE all highlight the difficulty of cross-generator and in-the-wild evaluation~\cite{zhu2023genimage,hong2024wildfake,pal2024semi,pellegrini2025ai,li2025artificial,xia2025mirage}, and Yan \emph{et al.} further show that realistic in-the-wild evaluation is substantially harder than standard cross-dataset testing~\cite{yan2024sanity}.

\noindent\textbf{Vision-language based and recent robust detectors.}
Pre-trained vision-language models have also been introduced for synthetic image detection. Built on the transferable representation of CLIP~\cite{radford2021learning}, CLIPDetection~\cite{cozzolino2024raising} demonstrates the value of semantic priors, while later works further improve this direction through language-guided contrastive learning, prompt-based adaptation, and semantic--pixel fusion~\cite{wu2025generalizable,tan2025c2p,cheng2025co}. However, semantic cues alone may become less reliable as recent generators produce increasingly coherent images. Recent robust detectors therefore approach the problem from complementary perspectives, including latent structural modeling~\cite{nguyen2026beyond}, native-resolution detail preservation~\cite{mu2025no}, real-boundary modeling~\cite{dhakal2026simlbr}, and spectral learning across resolutions~\cite{karageorgiou2025any}. Provenance-oriented methods such as diffusion watermarking have also been explored~\cite{wen2023tree}, but they rely on assumptions about the generation pipeline and thus differ from open-world forensic detection. Despite, most existing methods still follow a single-detector paradigm and do not explicitly model generator-oriented unique artifacts, leaving scalability an open challenge as new generators emerge.

\section{Method}
\subsection{LEGO Framework Overview}

We study AI-generated image detection under generator generalization. Given an input image $I$, the goal is to predict its authenticity label $y \in \{0,1\}$, where $y=0$ and $y=1$ denote real and fake images, respectively. Unlike conventional binary detectors that seek a single shared forensic cue across all generators, LEGO is motivated by the observation that as generator diversity increases, the common artifact region in feature space becomes increasingly limited, making universal traces less reliable for robust detection~\cite{park2025community,li2025artificial}. Instead, LEGO focuses on learning generator-specific artifact features.

Based on this insight, we propose {LEGO}, a generator-oriented modular detector built on a frozen CLIP ViT-L/14 visual encoder~\cite{radford2021learning}. As shown in Figure~\ref{fig:framework}, LEGO adopts a two-stage design. In the first stage, each LoRA block is trained on a specific generator family to capture its distinctive artifact patterns. In the second stage, these specialized LoRA blocks are frozen and assembled into a unified LoRA Hub. A shared MLP router predicts LoRA merge weights from the input feature representation, producing an input-adaptive integrated LoRA that is injected back into the frozen backbone before the final forgery head performs prediction. In this way, LEGO decomposes detection into LoRA specialization and composition, making the framework naturally aligned with generator diversity.

This design offers two key benefits. First, it avoids forcing incompatible artifact distributions into a single residual direction. Second, it decouples expert learning from expert coordination, allowing the model to first learn stable generator-specific priors and then adaptively combine them. As a result, LEGO models the synthetic artifact space as a set of generator-aware low-rank subspaces with an input-conditioned composition rule.
\subsection{Generator-Specific LoRA Specialization}
Each training sample is represented as $(I_i, y_i, g_i)$, where $I_i$ is the input image, $y_i \in \{0,1\}$ is the authenticity label, and $g_i \in \{1,\dots,K\}$ denotes the generator family. 
In our implementation, the training data are grouped by generator family, with additional real images included as authentic samples. When confronted with emerging generators, incorporating LoRA blocks~\cite{hu2022lora} and training a lightweight MLP router are much more efficient than designing an entirely new detection architecture, which means that the LEGO framework is naturally extensible to newly added generators.

Before being fed into the backbone, each image passes through the Data Augmentation \& Token Embedding stage:
\begin{equation}
I_i'=\mathcal{T}_{\mathrm{norm}}\big(\mathcal{T}_{\mathrm{aug}}(I_i)\big),
\end{equation}
where $\mathcal{T}_{\mathrm{aug}}$ includes horizontal flipping, JPEG compression, Gaussian blur, brightness/contrast adjustment, hue-saturation perturbation, and Gaussian noise, while $\mathcal{T}_{\mathrm{norm}}$ denotes resizing, tensor conversion, and CLIP-style normalization. This stage serves two purposes simultaneously: it aligns the input statistics with the frozen CLIP encoder~\cite{radford2021learning} and prevents each LoRA from overfitting to brittle low-level signals that are not stable under realistic post-processing~\cite{li2025improving,cavia2024real}. The preprocessed image is then converted into visual tokens using the standard CLIP token embedding scheme. 

Let $W_0 \in \mathbb{R}^{d_{\mathrm{in}}\times d_{\mathrm{out}}}$ denote a frozen projection matrix in the self-attention layer of CLIP. For each generator family, we introduce one generator-specific LoRA~\cite{hu2022lora} branch into the corresponding self-attention projection of every CLIP layer:
\begin{equation}
\Delta W_k = A_k B_k,\qquad k=1,\dots,K,
\end{equation}
where $A_k \in \mathbb{R}^{d_{\mathrm{in}}\times r}$, $B_k \in \mathbb{R}^{r\times d_{\mathrm{out}}}$, and $r \ll \min(d_{\mathrm{in}}, d_{\mathrm{out}})$ is the LoRA rank. The corresponding projection for the $k$-th LoRA is:
\begin{equation}
H_k(X)=XW_0+\frac{\alpha}{r}XA_kB_k,
\end{equation}
where $X \in \mathbb{R}^{N\times d_{\mathrm{in}}}$ denotes the token sequence of a transformer layer and $\alpha$ is the LoRA scaling factor.

This low-rank parameterization is central to the mathematical logic of LEGO. Rather than relearning the full projection matrix, each LoRA block is restricted to a compact residual subspace. Such a design is advantageous because the frozen CLIP backbone already encodes strong general visual semantics~\cite{radford2021learning}, while the missing information for forgery detection mainly lies in a comparatively small set of artifact-sensitive directions. Therefore, representing generator-specific knowledge as low-rank residuals is both parameter-efficient and structurally appropriate.

In Generator-Specific LoRA Specialization, only one LoRA branch is activated at a time, while the frozen backbone, all other LoRA branches, and the shared router remain fixed. For the $k$-th generator family, the model is optimized only on its corresponding subset. This design ensures that the gradient signal received by $(A_k,B_k)$ originates solely from one generator family, thereby avoiding destructive interference among heterogeneous generators and allowing each LoRA branch to focus on learning generator-specific artifact patterns. In this stage, the optimization is driven only by the authenticity classification loss:
\begin{equation}
	\mathcal{L}_{\mathrm{cls}}
	=
	-\frac{1}{B}\sum_{i=1}^{B}\sum_{c=0}^{1}\mathbbm{1}(y_i=c)\log p_{i,c},
\end{equation}
where $p_{i,c}$ denotes the predicted probability that the $i$-th sample belongs to class $c$. 

Under this training scheme, each LoRA branch can be interpreted as a generator-specific residual basis that approximates one local artifact subspace. This is precisely why we term this stage Generator-Specific LoRA Specialization: each LoRA is optimized to preserve the discriminative forgery traces of one generator family before any cross-generator composition is introduced.

\subsection{Generator-Oriented LoRA Composer}

Once the generator-specific LoRA blocks have been learned, we organize them into a LoRA Hub, as shown in Figure~\ref{fig:framework}. The Generator-Oriented LoRA Composer, implemented as a shared router, is designed to combine this set of discrete LoRA blocks into an input-adaptive Integrated LoRA and inject it back into the frozen backbone. In the following, we use {Composer} to refer to the overall composition module and {router} to emphasize its weight-prediction.

Given token features $X \in \mathbb{R}^{N\times d_{\mathrm{in}}}$ from the previous transformer block, we first compute a pooled descriptor:
\begin{equation}
	\bar{x}=\frac{1}{N}\sum_{n=1}^{N}X_n.
\end{equation}
This descriptor summarizes the current image at the feature level and serves as the input to the Composer. The shared router then predicts the LoRA merge weight:
\begin{equation}
	\pi(\bar{x})=\mathrm{Softmax}(\mathrm{MLP}(\bar{x})),
\end{equation}
where $\pi(\bar{x}) \in \mathbb{R}^{K}$ and $\sum_{k=1}^{K}\pi_k(\bar{x})=1$.

This formulation is mathematically meaningful for several reasons. First, the softmax constrains the routing weights to the probability simplex, which makes them directly interpretable as a convex combination over LoRA blocks. Second, the convexity constraint stabilizes optimization by preventing arbitrarily large contributions from a single branch. Third, because the router predicts continuous weights rather than hard assignments, the composer can smoothly interpolate between generators instead of being forced into an all-or-nothing decision. This is especially important for realistic test samples, whose traces may be ambiguous, mixed, or only partially aligned with previously seen generators.

The resulting Integrated LoRA is defined as:
\begin{equation}
H(X)=XW_0+\frac{\alpha}{r}\sum_{k=1}^{K}\pi_k(\bar{x})\,XA_kB_k.
\end{equation}
This equation makes the logic of the composer explicit. The first term, $XW_0$, preserves the generic visual representation learned by CLIP~\cite{radford2021learning}. The second term adds an input-dependent residual correction assembled from the LoRA Hub. Therefore, the model does not attempt to replace the backbone; instead, it learns to minimally and selectively perturb the backbone in artifact-sensitive directions that are most relevant to the current input.

From a geometric perspective, the set of generator-specific residuals $\{XA_kB_k\}_{k=1}^{K}$ spans a generator-aware residual space, and the composer determines a point in this space through $\pi(\bar{x})$. As a result, the final representation is not restricted to a single LoRA block, but is allowed to lie between multiple generator-specific subspaces. This property is one of the key reasons why LEGO remains robust when encountering unseen or weakly related generators~\cite{park2025community,li2025artificial}.

The composed feature is then passed to the Forgery Detection Head, which contains an Attention module followed by a Classification Head. Denoting the CLIP encoder equipped with the integrated LoRA by $\mathcal{E}_{\mathrm{LEGO}}(\cdot)$, the final prediction is:
\begin{equation}
p_i=\mathrm{Softmax}\!\left(W_c\,\mathrm{Attn}\!\left(\mathcal{E}_{\mathrm{LEGO}}(I_i')\right)+b_c\right),
\end{equation}
where $W_c$ and $b_c$ are the classifier parameters. The Detection Loss in Figure~\ref{fig:framework} is precisely the classification loss $\mathcal{L}_{\mathrm{cls}}$, which supervises the final real/fake decision.

\subsection{Generator-Aware Router Alignment}

After generator-specific LoRA specialization, we enter the second stage, namely Generator-Aware Router Alignment. In this stage, all generator-specific LoRA blocks are frozen, and only the shared router and the Forgery Detection Head are optimized. The objective is no longer to learn new generator-specific residuals, but to align the router with the learned residual basis so that it can produce meaningful LoRA Merge Weights for mixed-generator inputs.

To make the router aware of cross-generator ambiguity, we construct a GenMix Dataset by introducing fake-fake interpolation, inspired by mixup~\cite{zhang2017mixup}. Given two fake samples $I_a$ and $I_b$ from different generator families, we synthesize a harder example through:
\begin{equation}
\tilde{I}=\lambda I_a+(1-\lambda)I_b,
\end{equation}
where $\lambda \in [0,1]$ is the Random Mixing Ratio $w$. In Figure~\ref{fig:framework}, this ratio is also used as the supervisory coefficient for the routing branch. Since the image itself is a mixture of two generators, it is logically consistent to define its routing target as a mixture of the corresponding generator identities:
\begin{equation}
q_i=\lambda e_a+(1-\lambda)e_b,
\end{equation}
where $e_a$ and $e_b$ are one-hot generator labels. This yields the Generator Target Distribution shown in the figure. For ordinary non-fused samples, the target reduces to:
\begin{equation}
q_i=e_{g_i}.
\end{equation}

\begin{table*}[t]
	\centering
	\caption{Per-dataset performance comparison on AIGIBench (Part I, II, and III). The best and second-best results in each column are shown in bold and underlined, respectively.}
	\vspace{-8pt}
	\label{tab:aigibench}
	\setlength{\tabcolsep}{3.8pt}
	\renewcommand{\arraystretch}{1.08}
	\resizebox{\textwidth}{!}{
		\begin{tabular}{l|c|cc|cc|cc|cc|cc|cc|cc|cc}
			
			\toprule
			\multicolumn{1}{c|}{\textbf{Test Dataset} $\rightarrow$} 
			& \multirow{2}{*}{\textbf{Venue}}
			& \multicolumn{2}{c|}{\textbf{R3GAN}}
			& \multicolumn{2}{c|}{\textbf{StyleGAN3}}
			& \multicolumn{2}{c|}{\textbf{StyleGAN-XL}}
			& \multicolumn{2}{c|}{\textbf{StyleSwim}}
			& \multicolumn{2}{c|}{\textbf{WFIR}}
			& \multicolumn{2}{c|}{\textbf{BlendFace}}
			& \multicolumn{2}{c|}{\textbf{E4S}}
			& \multicolumn{2}{c}{\textbf{FaceSwap}} \\
			\cmidrule(r){1-1}\cmidrule(lr){3-18}
			\multicolumn{1}{c|}{\textbf{Detectors} $\downarrow$}
			&
			& Acc. & A.P.
			& Acc. & A.P.
			& Acc. & A.P.
			& Acc. & A.P.
			& Acc. & A.P.
			& Acc. & A.P.
			& Acc. & A.P.
			& Acc. & A.P. \\
			\midrule
			
			ResNet-50    & CVPR'16  & 48.5 & 53.7 & 70.8 & 88.3 & 60.0 & 76.7 & 83.2 & 93.7 & 49.5 & 56.8 & 46.4 & 34.3 & 46.7 & 32.9 & 48.8 & 39.8 \\
			CNNDetection~\cite{wang2020cnn}   & CVPR'20  & 50.4 & 52.7 & 55.8 & 73.1 & 52.8 & 64.2 & 52.6 & 76.5 & 49.8 & 50.0 & \textbf{52.4} & \textbf{73.4} & 51.1 & \underline{68.9} & 50.3 & 58.7 \\
			Gram-net~\cite{liu2020global}      & CVPR'20  & 47.9 & 52.5 & 65.6 & 77.9 & 72.9 & 83.7 & 75.5 & 86.0 & 44.5 & 43.9 & 42.3 & 33.3 & 42.5 & 32.5 & 47.0 & 34.6 \\
			LGrad~\cite{tan2023learning}         & CVPR'23  & 54.4 & 58.7 & 70.5 & 80.5 & 65.7 & 74.6 & 81.3 & 90.0 & 51.7 & 49.4 & 41.8 & 34.9 & 41.5 & 32.8 & 45.3 & 37.5 \\
			CLIPDetection~\cite{cozzolino2024raising} & CVPR'23  & 83.5 & 91.2 & 79.6 & 84.5 & 84.6 & 93.3 & 86.4 & 95.2 & 70.0 & 82.0 & 35.0 & 35.3 & 57.0 & 57.1 & \underline{53.1} & 52.4 \\
			FreqNet~\cite{tan2024frequency}       & AAAI'23  & 62.3 & 56.8 & 83.0 & 92.4 & 79.8 & 84.1 & 80.8 & 91.8 & 58.5 & 48.9 & 23.3 & 34.1 & 25.8 & 34.7 & 40.4 & 43.4 \\
			NPR~\cite{tan2024rethinking}           & CVPR'24  & 50.8 & 61.1 & 78.4 & 91.7 & 60.3 & 75.3 & 85.7 & 94.9 & 51.6 & 65.5 & 44.5 & 34.7 & 45.0 & 34.4 & 48.1 & 43.6 \\
			DFFreq~\cite{yan2026dual}        & arXiv'25 & 61.7 & 74.0 & \underline{90.1} & 96.4 & 73.7 & 83.4 & 84.5 & 92.9 & 74.6 & 82.2 & 41.5 & 35.2 & 42.1 & 34.8 & 47.6 & 45.6 \\
			LaDeDa~\cite{cavia2024real}        & arXiv'24 & 54.8 & 72.6 & \textbf{92.4} & \underline{96.9} & \textbf{94.7} & \textbf{98.5} & 94.0 & \underline{98.5} & 58.5 & 86.9 & 42.4 & 42.1 & 42.9 & 49.3 & 47.1 & 40.9 \\
			AIDE~\cite{yan2024sanity}          & ICLR'25  & \textbf{92.9} & \textbf{97.1} & 88.1 & 91.4 & 88.7 & 93.2 & 83.7 & 89.3 & 71.4 & 90.8 & 51.5 & 54.2 & 44.3 & 44.3 & 52.1 & 56.3 \\
			CO-SPY~\cite{cheng2025co}        & CVPR'25  & 84.5 & 95.9 & 84.5 & 91.3 & 85.4 & \underline{96.1} & 79.6 & 93.4 & 74.8 & 87.4 & 41.3 & 41.7 & \underline{68.0} & 67.4 & 50.0 & 50.3 \\
			Effort~\cite{yan2024effort}        & ICML'25  & 90.2 & 95.8 & 87.9 & 94.9 & \underline{92.6} & 93.7 & \textbf{94.5} & \textbf{99.7} & \underline{85.5} & \underline{93.8} & 41.4 & 40.4 & 49.6 & 56.7 & 48.8 & \underline{59.9} \\
			\rowcolor{gray!18}
			Ours          & -        & \underline{90.8} & \underline{96.7} & 87.9 & \textbf{97.1} & 65.5 & 80.4 & \underline{94.1} & 98.4 & \textbf{86.1} & \textbf{95.8} & \underline{51.9} & \underline{59.4} & \textbf{76.8} & \textbf{92.0} & \textbf{68.1} & \textbf{82.8} \\
			
		\end{tabular}
	}
\end{table*}

\begin{table*}[t]
	\centering
	\label{tab:aigibench_part2_no_safe_no_progan}
	\scriptsize
	\setlength{\tabcolsep}{3.8pt}
	\renewcommand{\arraystretch}{1.08}
	\resizebox{\textwidth}{!}{
		\begin{tabular}{l|c|cc|cc|cc|cc|cc|cc|cc|cc}
			\toprule
			\multicolumn{1}{c|}{\textbf{Test Dataset} $\rightarrow$} 
			& \multirow{2}{*}{\textbf{Venue}}
			& \multicolumn{2}{c|}{\textbf{InSwap}}
			& \multicolumn{2}{c|}{\textbf{SimSwap}}
			& \multicolumn{2}{c|}{\textbf{FLUX1-dev}}
			& \multicolumn{2}{c|}{\textbf{Midjourney}}
			& \multicolumn{2}{c|}{\textbf{GLIDE}}
			& \multicolumn{2}{c|}{\textbf{DALLE-3}}
			& \multicolumn{2}{c|}{\textbf{Imagen3}}
			& \multicolumn{2}{c}{\textbf{SD3}} \\
			\cmidrule(r){1-1}\cmidrule(lr){3-18}
			\multicolumn{1}{c|}{\textbf{Detectors} $\downarrow$}
			&
			& Acc. & A.P.
			& Acc. & A.P.
			& Acc. & A.P.
			& Acc. & A.P.
			& Acc. & A.P.
			& Acc. & A.P.
			& Acc. & A.P.
			& Acc. & A.P. \\
			\midrule
			ResNet-50     & CVPR'16  & 48.8 & 39.7 & 48.1 & 40.3 & 82.3 & 93.6 & 52.5 & 57.4 & 79.3 & 93.4 & 53.7 & 68.4 & 61.9 & 78.6 & 64.2 & 83.7 \\
			CNNDetection~\cite{wang2020cnn}  & CVPR'20  & \underline{54.5} & \textbf{77.9} & 52.1 & \textbf{70.0} & 57.4 & 72.3 & 52.3 & 59.8 & 51.1 & 60.0 & 53.9 & 68.6 & 51.2 & 57.4 & 55.8 & 73.1 \\
			Gram-net~\cite{liu2020global}      & CVPR'20  & 47.1 & 38.0 & 46.7 & 36.4 & 64.5 & 75.1 & 43.9 & 41.6 & 71.6 & 84.3 & 53.5 & 61.1 & 50.6 & 54.7 & 52.5 & 60.0 \\
			LGrad~\cite{tan2023learning}         & CVPR'23  & 44.6 & 35.0 & 44.2 & 37.6 & 80.4 & 88.1 & 60.4 & 64.5 & 82.4 & 91.0 & 57.2 & 62.7 & 62.2 & 69.7 & 63.4 & 72.1 \\
			CLIPDetection~\cite{cozzolino2024raising} & CVPR'23  & 43.7 & 40.2 & 43.7 & 40.4 & 80.0 & 79.5 & 65.3 & 61.5 & 76.7 & 80.3 & 75.1 & 76.3 & 78.9 & 79.3 & 84.5 & 87.2 \\
			FreqNet~\cite{tan2024frequency}       & AAAI'23  & 37.5 & 42.1 & 36.5 & 41.9 & 78.5 & 87.3 & 53.9 & 55.9 & 75.8 & 77.4 & 66.2 & 61.0 & 73.6 & 80.7 & 77.3 & 82.6 \\
			NPR~\cite{tan2024rethinking}           & CVPR'24  & 47.8 & 40.7 & 47.4 & 42.7 & \textbf{95.2} & \textbf{99.0} & 68.8 & 76.9 & 82.5 & 94.3 & 57.1 & 70.0 & 85.9 & 94.4 & 91.9 & 97.2 \\
			DFFreq~\cite{yan2026dual}        & arXiv'25 & 47.0 & 41.2 & 45.6 & 43.8 & 76.6 & 86.3 & 64.6 & 68.1 & \underline{88.5} & \underline{96.3} & 51.8 & 58.6 & 75.7 & 87.3 & 81.3 & 90.8 \\
			LaDeDa~\cite{cavia2024real}        & arXiv'24 & 47.0 & 47.4 & 46.3 & 42.3 & 94.6 & 98.7 & \underline{80.3} & \underline{86.9} & 87.1 & 95.0 & 50.1 & 59.8 & \underline{91.6} & \underline{97.2} & \underline{95.1} & \underline{98.7} \\
			AIDE~\cite{yan2024sanity}          & ICLR'25  & 50.9 & 54.6 & \underline{54.9} & 62.7 & 88.0 & 93.4 & 76.4 & 83.0 & \textbf{93.4} & \textbf{97.7} & 55.1 & 63.1 & 89.8 & 95.2 & 94.3 & 98.3 \\
			CO-SPY~\cite{cheng2025co}        & CVPR'25  & 36.4 & 41.4 & 39.8 & 41.0 & 77.7 & 87.3 & 69.4 & 74.6 & 75.2 & 84.6 & \underline{81.1} & \underline{85.6} & 79.1 & 76.9 & 89.3 & 95.6 \\
			Effort~\cite{yan2024effort}        & ICML'25  & 48.0 & 51.3 & 52.3 & 58.4 & 75.4 & 81.2 & 73.8 & 77.4 & 83.6 & 93.5 & 77.0 & 82.7 & 65.6 & 74.7 & 82.4 & 91.9 \\
			\rowcolor{gray!18}
			Ours          & -        & \textbf{55.4} & \underline{65.5} & \textbf{58.2} & \underline{69.2} & \underline{95.1} & \underline{98.9} & \textbf{98.6} & \textbf{99.9} & 77.7 & 91.7 & \textbf{97.6} & \textbf{99.9} & \textbf{96.2} & \textbf{99.2} & \textbf{97.4} & \textbf{99.8} \\
			
		\end{tabular}
	}
\end{table*}

\begin{table*}[t]
	\centering
	\label{tab:aigibench_part3_no_safe_no_progan}
	\scriptsize
	\setlength{\tabcolsep}{3.8pt}
	\renewcommand{\arraystretch}{1.08}
	\resizebox{\textwidth}{!}{
		\begin{tabular}{l|c|cc|cc|cc|cc|cc|cc|cc|cc|cc}
			\toprule
			\multicolumn{1}{c|}{\textbf{Test Dataset} $\rightarrow$} 
			& \multirow{2}{*}{\textbf{Venue}}
			& \multicolumn{2}{c|}{\textbf{SDXL}}
			& \multicolumn{2}{c|}{\textbf{BLIP}}
			& \multicolumn{2}{c|}{\textbf{Infinite-ID}}
			& \multicolumn{2}{c|}{\textbf{InstantID}}
			& \multicolumn{2}{c|}{\textbf{IP-Adapter}}
			& \multicolumn{2}{c|}{\textbf{PhotoMaker}}
			& \multicolumn{2}{c|}{\textbf{SocialRF}}
			& \multicolumn{2}{c|}{\textbf{CommunityAI}}
			& \multicolumn{2}{c}{\textbf{Mean}} \\
			\cmidrule(r){1-1}\cmidrule(lr){3-20}
			\multicolumn{1}{c|}{\textbf{Detectors} $\downarrow$}
			&
			& Acc. & A.P.
			& Acc. & A.P.
			& Acc. & A.P.
			& Acc. & A.P.
			& Acc. & A.P.
			& Acc. & A.P.
			& Acc. & A.P.
			& Acc. & A.P.
			& Acc. & A.P. \\
			\midrule
			ResNet-50     & CVPR'16  & 72.8 & 89.8 & \textbf{99.5} & \textbf{100.0} & 49.9 & 62.6 & 61.0 & 80.7 & 62.8 & 83.1 & 49.1 & 58.7 & 55.3 & 61.5 & 52.5 & 64.2 & 60.3 & 68.0 \\
			CNNDetection~\cite{wang2020cnn}  & CVPR'20  & 52.8 & 64.2 & 77.2 & 92.9 & 49.7 & 49.5 & 53.2 & 80.2 & 52.0 & 65.8 & 50.1 & 58.2 & 51.1 & 50.6 & 51.3 & 59.1 & 53.4 & 65.7 \\
			Gram-net~\cite{liu2020global}      & CVPR'20  & 63.8 & 76.9 & 98.6 & \underline{99.9} & 50.6 & 60.1 & 75.1 & 85.7 & 54.5 & 64.2 & 50.1 & 58.6 & 52.1 & 53.0 & 52.6 & 66.1 & 56.9 & 60.8 \\
			LGrad~\cite{tan2023learning}         & CVPR'23  & 73.6 & 82.8 & 93.0 & 97.4 & 50.9 & 54.6 & 72.6 & 81.5 & 70.3 & 78.3 & 59.9 & 67.2 & 53.5 & 54.9 & 55.5 & 69.4 & 61.5 & 65.2 \\
			CLIPDetection~\cite{cozzolino2024raising} & CVPR'23  & 84.7 & 88.0 & 88.6 & 95.8 & 84.5 & 89.6 & 85.4 & 93.5 & 82.6 & 87.3 & 69.3 & 72.3 & 54.4 & 55.2 & \underline{67.0} & \underline{73.2} & 71.4 & 74.6 \\
			FreqNet~\cite{tan2024frequency}       & AAAI'23  & 82.7 & 95.2 & 93.8 & \textbf{100.0} & 79.0 & 74.5 & 79.8 & 86.3 & 78.8 & 79.9 & 77.0 & 74.9 & 54.2 & 58.1 & 55.9 & 69.7 & 64.8 & 68.9 \\
			NPR~\cite{tan2024rethinking}           & CVPR'24  & 86.6 & 94.4 & \underline{99.2} & \textbf{100.0} & 63.9 & 80.4 & 63.8 & 79.2 & 82.4 & 91.7 & 48.1 & 43.6 & 59.1 & 68.4 & 54.0 & 62.9 & 66.6 & 72.4 \\
			DFFreq~\cite{yan2026dual}        & arXiv'25 & 88.9 & 95.8 & 97.9 & 99.6 & 70.4 & 82.7 & \underline{91.9} & \underline{97.2} & 83.2 & 91.3 & \underline{88.0} & \underline{94.0} & 57.6 & 63.3 & 54.5 & 52.1 & 70.0 & 74.7 \\
			LaDeDa~\cite{cavia2024real}        & arXiv'24 & \underline{94.7} & \underline{98.5} & 99.0 & \underline{99.9} & 61.5 & 76.9 & 86.5 & 90.4 & \textbf{90.8} & \underline{94.3} & 78.4 & 90.7 & 58.6 & 68.3 & 54.5 & 56.3 & 72.6 & 78.6 \\
			AIDE~\cite{yan2024sanity}          & ICLR'25  & 93.5 & 95.7 & 96.4 & 95.5 & \textbf{92.2} & 94.7 & 91.8 & 96.3 & \underline{90.0} & \textbf{95.4} & \textbf{91.7} & \textbf{95.6} & 57.8 & 65.0 & 54.1 & 61.0 & \underline{76.8} & \underline{81.8} \\
			CO-SPY~\cite{cheng2025co}        & CVPR'25  & 85.4 & 92.6 & 85.0 & 95.5 & \underline{89.8} & \underline{97.4} & 84.0 & 96.7 & 74.8 & 84.3 & 55.8 & 57.3 & \underline{61.7} & \underline{72.9} & 62.6 & 69.3 & 71.5 & 78.2 \\
			Effort~\cite{yan2024effort}        & ICML'25  & 79.3 & 91.9 & 98.0 & \underline{99.9} & \textbf{92.2} & \textbf{98.7} & \textbf{92.2} & \textbf{99.0} & 87.5 & 91.0 & 76.6 & 82.3 & 53.1 & 57.1 & 51.6 & 60.1 & 74.1 & 80.2 \\
			\rowcolor{gray!18}
			Ours          & -        & \textbf{97.6} & \textbf{99.8} & 73.3 & 89.7 & 86.7 & 96.0 & 89.2 & 97.1 & 64.8 & 83.0 & 62.2 & 80.5 & \textbf{68.1} & \textbf{83.9} & \textbf{85.4} & \textbf{93.1} & \textbf{80.2} & \textbf{89.6} \\
			\bottomrule
		\end{tabular}
	}
\end{table*}

Let $\hat{q}_i=\pi(\bar{x}_i)$ denote the predicted routing distribution. We supervise the router with the Generator-Aware Routing Loss:
\begin{equation}
\mathcal{L}_{\mathrm{route}}
=
-\frac{1}{B}\sum_{i=1}^{B}\sum_{k=1}^{K}q_{i,k}\log \hat{q}_{i,k}.
\end{equation}
This loss is mathematically aligned with the design of the composer. Since the router outputs a probability simplex and the target distribution is defined on the same simplex, the soft cross-entropy encourages the predicted LoRA merge weight to match the LoRA composition. Compared with hard one-hot supervision, this soft target preserves uncertainty and teaches the router that generator membership may be continuous rather than discrete.

We additionally apply an $L_1$ regularization term to the router:
\begin{equation}
\mathcal{L}_{\mathrm{reg}}=\sum_{\theta\in\Theta_{\mathrm{router}}}\|\theta\|_1.
\end{equation}
Its role is to prevent overly diffuse compositions and to encourage the router to rely on a compact subset of LoRAs. This sparsity bias is logically desirable: although the composer allows continuous interpolation, a given image is unlikely to require all generator priors simultaneously.

The full second-stage objective is thus:
\begin{equation}
\mathcal{L}_{\mathrm{stage2}}
=
\mathcal{L}_{\mathrm{cls}}
+
\lambda_{\mathrm{route}}\mathcal{L}_{\mathrm{route}}
+
\lambda_{\mathrm{reg}}\mathcal{L}_{\mathrm{reg}},
\end{equation}
where $\lambda_{\mathrm{route}}$ and $\lambda_{\mathrm{reg}}$ control the contributions of routing supervision and regularization, respectively.

This objective reveals the internal logic of Generator-Aware Router Alignment. The term $\mathcal{L}_{\mathrm{cls}}$ guarantees that the composed representation remains discriminative for authenticity prediction. The term $\mathcal{L}_{\mathrm{route}}$ aligns the router with the latent generator composition of the input. The term $\mathcal{L}_{\mathrm{reg}}$ keeps the composition sparse and stable. Together, they ensure that the second stage does not merely fit labels, but learns a structurally meaningful mapping from mixed-generator inputs to LoRA combinations.

Overall, LEGO can be understood as a two-stage approximation to the artifact manifold of AI-generated images. Generator-Specific LoRA Specialization learns the basis elements of that manifold, Generator-Oriented LoRA Composer constructs an input-conditioned residual from these basis elements, and Generator-Aware Router Alignment ensures that the composition rule is both discriminative and semantically consistent with generator identity. This makes the framework not only effective in practice, but also mathematically coherent and logically extensible.
\section{Experimental Results and Analysis}

\subsection{Experimental Setup}
\setlength{\dblfloatsep}{0.2pt}
\setlength{\dbltextfloatsep}{4pt}

\noindent\textbf{Training datasets.}To validate the effectiveness of LEGO under a practical low-cost training setting, we intentionally restrict the training data to only a few compact GenImage subsets~\cite{zhu2023genimage}. Specifically, we use only the held-out split of three representative generator families, namely Stable Diffusion v1.4 \cite{rombach2022high}, ProGAN\cite{karras2017progressive}, and ADM\cite{dhariwal2021diffusion}, resulting in fewer than 30,000 training images in total. Although highly compact, these subsets still cover both diffusion-based and GAN-based generators, and thus provide diverse supervision for generator-oriented artifact learning.

Notably, this training scale is dramatically smaller than those used in prior work. For instance, the ProGAN training split in GenImage~\cite{zhu2023genimage} alone already contains about $20 \times 18{,}000$ images, which far exceeds our total training data. Our setting is also considerably lighter than that of Effort~\cite{yan2024effort}, which trains on two GenImage subsets (SDv1.4 and ProGAN), and much more efficient than HiDA-Net~\cite{mu2025no}, which uses all available training subsets. During mixed training, samples from the selected subsets are randomly shuffled and jointly optimized. This setting therefore highlights a key advantage of LEGO: it achieves strong generalization with substantially lower training cost, making it more suitable for scalable adaptation to newly emerging generators.

\noindent\textbf{Test datasets.}To evaluate generalization to future unseen generators, we train on GenImage~\cite{zhu2023genimage} and test on AIGIBench and Chameleon~\cite{li2025artificial,yan2024sanity}, neither of which is used for end-to-end retraining. The results therefore reflect transfer from limited seen generators to more diverse real-world distributions.

For a fairer out-of-distribution comparison on AIGIBench~\cite{li2025artificial}, we exclude ProGAN, as it is already represented by the generator families covered in the training data. 

\noindent\textbf{Implementation details.}We implement LEGO based on a pre-trained OpenCLIP/CLIP vision backbone~\cite{radford2021learning} and train the model on a single NVIDIA RTX A6000 GPU. Unless otherwise specified, the detector is optimized with AdamW, where the learning rate is set to $2\times10^{-5}$, weight decay to $5\times10^{-4}$, $\beta_1=0.9$, $\beta_2=0.95$,  $\epsilon=10^{-8}$, and 5 epochs of training. The model is trained under the mixed-generator setting, where the three selected GenImage subsets~\cite{zhu2023genimage} are randomly interleaved during optimization. This protocol simulates a practical scenario in which only partial generator data are available for training, while the detector is still expected to generalize to unseen synthesis domains.

\begin{table}[t]
	\centering
	\caption{Comparison with existing methods on the Chameleon benchmark. Results for HiDA-Net are reported as provided in the original paper, where only ACC is available. The best and second-best results in each column are shown in bold and underlined, respectively.}
	\vspace{-5pt}
	\label{tab:chameleon_main}
	\setlength{\tabcolsep}{2.8pt}
	\renewcommand{\arraystretch}{1.12}
	\begin{tabular}{ll|cccc}
		\toprule
		Method & Venue & AUC $\uparrow$ & ACC $\uparrow$ & EER $\downarrow$ & A.P. $\uparrow$ \\
		\midrule
		CNNDetection~\cite{wang2020cnn}                 & CVPR 2020  & 63.3 & 58.4 & 40.6 & 58.0 \\
		FreDect~\cite{frank2020leveraging}         & ICML 2020  & 75.5 & 63.0 & 31.4 & 68.6 \\
		Fusing~\cite{ju2022fusing}                 & ICIP 2022  & 60.2 & 57.5 & 41.4 & 53.5 \\
		LGrad~\cite{tan2023learning}               & CVPR 2023  & 53.4 & 51.6 & 48.1 & 47.4 \\
		LNP~\cite{liu2022detecting}                & ECCV 2022  & 58.0 & 57.9 & 44.2 & 50.4 \\
		CORE~\cite{ni2022core}                     & CVPRW 2022 & 52.6 & 57.0 & 47.7 & 45.6 \\
		SPSL~\cite{liu2021spatial}                 & CVPR 2021  & 61.4 & 57.5 & 41.5 & 53.4 \\
		UIA-ViT~\cite{zhuang2022uia}               & ECCV 2022  & 70.4 & 59.9 & 35.4 & 61.2 \\
		DIRE~\cite{wang2023dire}                   & ICCV 2023  & 54.9 & 57.7 & 45.7 & 49.1 \\
		UnivFD~\cite{ojha2023towards}              & CVPR 2023  & 80.3 & 66.0 & 26.7 & 75.1 \\
		AIDE~\cite{yan2024sanity}                  & ICLR 2025  & 79.7 & 62.6 & 27.5 & 73.5 \\
		NPR~\cite{tan2024rethinking}               & CVPR 2024  & 58.7 & 58.1 & 43.8 & 49.4 \\
		Effort~\cite{yan2024effort}                & ICML 2025  & 83.7 & 64.6 & 24.3 & 76.9 \\
		TriDetect~\cite{nguyen2026beyond}          & AAAI 2026  & \underline{89.4} & 67.9 & \underline{18.4} & \underline{87.4} \\
		HiDA-Net$^\dagger$~\cite{mu2025no}         & ICLR 2026  & -- & \underline{79.1} & -- & -- \\
		\midrule
		LEGO                                       & Ours       & \textbf{92.8} & \textbf{83.6} & \textbf{15.0} & \textbf{89.9} \\
		\bottomrule
	\end{tabular}
		
	\begin{minipage}{0.98\linewidth}
		\footnotesize
		$^\dagger$ Reported under the All GenImage training setting.
		HiDA-Net only reports ACC on Chameleon in its original paper.
	\end{minipage}
\end{table}

\subsection{Comparison with SOTA Methods}
\noindent\textbf{Results on AIGIBench.}
Tables~\ref{tab:aigibench} compare LEGO with recent state-of-the-art detectors on AIGIBench~\cite{li2025artificial}. 
Overall, LEGO achieves the best average performance on AIGIBench, reaching 80.2\% mean ACC and 89.6\% mean AP. Among the compared baselines, AIDE~\cite{yan2024sanity} is the strongest overall with 76.8\% mean ACC and 81.8\% mean AP, while Effort~\cite{yan2024effort} achieves 74.1\% and 80.3\%, respectively. These results indicate that LEGO provides the strongest overall generalization across diverse generator families.

More importantly, LEGO exhibits clear advantages on several challenging subsets, particularly those involving deepfake-style manipulation and open-source generation platforms. For example, LEGO achieves the best accuracy on BlendFace, E4S, FaceSwap, InSwap, SimSwap, and CommunityAI. On the AP metric, LEGO further ranks first on R3GAN, WFIR, E4S, FaceSwap, InSwap, SimSwap, and CommunityAI. These gains are particularly important because such subsets usually contain more diverse and less semantically obvious forgery traces, where detectors relying mainly on shared artifacts often degrade significantly~\cite{ojha2023towards,park2025community,zheng2024breaking}.

\begin{table*}[t]
	\centering
	\caption{Ablation on core design choices.}
	\vspace{-7pt}
	\label{tab:ab_core}
	\small
	\setlength{\tabcolsep}{8pt}
	\renewcommand{\arraystretch}{1.05}
	\begin{tabular}{l|cccc|cc|cc}
		\toprule
		\multirow{2}{*}{Method} 
		& \multirow{2}{*}{Multi-LoRA} 
		& \multirow{2}{*}{Router} 
		& \multirow{2}{*}{Route Loss} 
		& \multirow{2}{*}{Two-Stage}
		& \multicolumn{2}{c|}{AIGIBench}
		& \multicolumn{2}{c}{Chameleon} \\
		\cmidrule(lr){6-7} \cmidrule(l){8-9}
		& & & & & Acc & A.P. & Acc & A.P. \\
		\midrule
		CLIP + Single LoRA      &  &  &  &  & 67.0 & 74.4 & 59.0 & 64.7 \\
		+ Multi-branch LoRA Hub & \checkmark &  &  &  & 75.9 & 85.0 & 58.6 & 79.4 \\
		+ Shared Router         & \checkmark & \checkmark &  &  &70.6 & 84.5 & 57.6 & 73.6 \\
		+ Routing Supervision   & \checkmark & \checkmark & \checkmark &  & 68.4 & 76.2 & 60.0 & 83.5 \\
		LEGO (Full)             & \checkmark & \checkmark & \checkmark & \checkmark & \textbf{80.2} & \textbf{89.6} & \textbf{83.6} & \textbf{89.9} \\
		\bottomrule
	\end{tabular}
\end{table*}

We also observe that several recent detectors, such as AIDE~\cite{yan2024sanity}, can achieve strong performance on specific generator categories, especially diffusion-based subsets. However, their advantages are often concentrated on a limited group of generators, and their performance fluctuates more noticeably across different synthesis paradigms. By contrast, LEGO shows stronger overall generalization across GAN-based, diffusion-based, personalized-generation, and deepfake-style subsets. This result is consistent with our main motivation: rather than forcing all forgery traces into a single universal representation, LEGO improves generalization by explicitly modeling generator-specific artifact patterns and adaptively composing them during inference.

It is also worth noting that LEGO achieves these results under a substantially more economical training setting. Therefore, the superiority of LEGO is not merely due to larger-scale training, but rather comes from a more effective generator-oriented decomposition of artifact space. This further validates the proposed design as a scalable and practical solution for open-world AI-generated image detection~\cite{park2025community,li2025artificial}.

\noindent\textbf{Results on Chameleon.}
In-the-wild benchmarks provide a more realistic and challenging test bed for generalized AI-generated image detection, since they contain perceptually deceptive samples with substantially broader semantic and visual variations than conventional cross-generator benchmarks. We therefore further evaluate LEGO on the challenging Chameleon benchmark.

As shown in Table~\ref{tab:chameleon_main}, LEGO achieves the best performance on all reported metrics, reaching 92.8 AUC, 83.6 ACC, 15.0 EER, and 89.9 AP. Compared with TriDetect, LEGO improves AUC/ACC/AP by 3.4/15.7/2.5 percentage points, respectively, and reduces EER by 3.4 points. We also include the recent 2026 method HiDA-Net, which reports Chameleon performance only in terms of ACC under its own training setting. Even under this reported setting, LEGO still surpasses HiDA-Net by 4.5 percentage points in ACC.

Overall, these results show that LEGO captures more transferable forensic cues and generalizes better to human-hard, in-the-wild synthetic images, where existing detectors often suffer substantial performance degradation.

\begin{table}[t]
	\centering
	\caption{Ablation on training and data strategy.}
	\vspace{-7pt}
	\label{tab:ab_data}
	\small
	\setlength{\tabcolsep}{4pt}
	\renewcommand{\arraystretch}{1.05}
	\begin{tabular}{l|cc|cc}
		\toprule
		Method 
		& \multicolumn{2}{c|}{AIGIBench}
		& \multicolumn{2}{c}{Chameleon} \\
		\cmidrule(lr){2-3} \cmidrule(lr){4-5}
		& Acc & A.P. & Acc & A.P. \\
		\midrule
		LEGO w/o both              & 77.6 & 85.3 & 66.2 & 78.3 \\
		LEGO w/o augmentation      & 76.8 & 84.5 & 66.7 & 75.2 \\
		LEGO w/o fake--fake fusion & 61.7 & 78.8 & 72.5 & \textbf{91.1} \\
		LEGO (full)                & \textbf{80.2} & \textbf{89.6} & \textbf{83.6} & {89.9} \\
		\bottomrule
	\end{tabular}
	\vspace{-7pt}
\end{table}

\begin{table}[t]
	\centering
	\caption{Effect of the number of LoRA branches $K$.}
	\vspace{-7pt}
	\label{tab:ab_k}
	\small
	\setlength{\tabcolsep}{10pt}
	\renewcommand{\arraystretch}{1.05}
	\begin{tabular}{c|cc|cc}
		\toprule
		$K$
		& \multicolumn{2}{c|}{AIGIBench}
		& \multicolumn{2}{c}{Chameleon} \\
		\cmidrule(lr){2-3} \cmidrule(lr){4-5}
		& Acc & A.P. & Acc & A.P. \\
		\midrule
		1 & 67.0 & 74.4 & 59.0 & 64.7 \\
		2 & 76.2 & 83.6 & 80.0 & 87.8 \\
		3 & 80.2 & {89.6} & {83.6} & {89.9} \\
		4 & 82.4 & 90.1 & 85.0 & 92.4 \\
		\bottomrule
	\end{tabular}
	\vspace{-7pt}
\end{table}

\subsection{Ablation Study}

We conduct ablation studies to analyze the contribution of each design choice in LEGO under the same low-cost training setting as the main experiments. Unless otherwise specified, all variants are trained with the same backbone and evaluated on both AIGIBench and Chameleon. We mainly report detection accuracy (\%) for compact presentation.

\noindent\textbf{Ablation on Individual Modules.} We first study the contribution of the main architectural components, including the generator-oriented multi-branch LoRA hub, the shared router, the generator-aware routing loss, and the proposed two-stage optimization. Starting from a basic CLIP+single-LoRA baseline, we progressively add each component.

Table~\ref{tab:ab_core} shows that simply increasing the number of LoRA branches does not consistently translate to better generalization: while the multi-branch LoRA hub improves AIGIBench and Chameleon AP, it does not improve Chameleon ACC. Adding a shared router without routing supervision further destabilizes performance, especially on AIGIBench. Routing supervision partially recovers the benefits, but the full two-stage design is the only variant that consistently improves all metrics on both benchmarks.

\noindent\textbf{Effect of Training and Data Strategy.}We next investigate the impact of the training-time data pipeline. In particular, our implementation adopts CLIP-compatible preprocessing together with image-level augmentation and a low-probability fake--fake fusion strategy, where fake images from different generator families are linearly blended during mixed training. We compare the full model against variants without augmentation or without cross-generator fusion. 
Although removing fake--fake fusion can improve a specific metric on a specific benchmark (e.g., AP on Chameleon), it substantially reduces the overall balance of generalization, especially on the more diverse multi-source benchmark AIGIBench.
Accordingly, its advantage is more clearly reflected on diverse multi-source benchmarks such as AIGIBench, where it yields better overall generalization performance.

As shown in Table~\ref{tab:ab_data}, both training-time augmentation and cross-generator fusion improve robustness. The augmentation pipeline prevents overfitting to low-level fingerprints, while fake--fake fusion encourages smoother router behavior under cross-generator ambiguity. Their combination gives the strongest generalization.

\noindent\textbf{Effect of the Number of LoRA Branches.}Finally, we study the influence of the number of generator-oriented LoRA branches $K$. In principle, using more LoRA branches provides a finer decomposition of the synthetic artifact space and thus leads to better performance. However, increasing $K$ also introduces higher training and optimization cost. Considering the trade-off between performance and efficiency, we adopt $K=3$ as the default setting.

Table~\ref{tab:ab_k} shows that increasing the number of LoRA branches $K$ consistently improves performance, confirming the benefit of decomposing artifact learning into multiple generator-aware branches. In principle, a larger $K$ allows a finer partition of the synthetic artifact space and yields stronger detection capability. However, it also introduces higher training cost and model complexity. Therefore, considering the trade-off between effectiveness and efficiency, we adopt $K=3$ as the default setting in our experiments.

Overall, the ablation results consistently support our design choices. The gain of LEGO does not come from any single factor, but from the synergy of generator-specific LoRA specialization, adaptive routing, robustness-oriented data processing, and the proposed two-stage training scheme.

\section{Conclusion}

Overall, the experimental results verify three major strengths of LEGO. First, the proposed framework achieves strong cross-dataset generalization on AIGIBench and Chameleon. Second, LEGO remains effective even when trained on only a subset of GenImage, demonstrating its efficiency under limited-resource settings. Third, its modular LoRA-based design offers a scalable path toward future adaptation: as next-generation generators emerge, new LoRA modules can be added in a plug-and-play manner, much like assembling LEGO bricks.
These findings support our central claim that, for modern AIGC detection, explicitly modeling generator-oriented unique artifacts is more effective than relying solely on a single universal artifact space.

\bibliographystyle{ACM-Reference-Format}
\bibliography{sample-base}


\end{document}